\documentclass[journal]{IEEEtran}
\usepackage{cite}
\usepackage{xspace} 
\usepackage{amsmath,amssymb,amsfonts} 
\usepackage{amsthm} 
\usepackage{hyperref} 
\usepackage{graphics} 
\graphicspath{{figs/}}
 \usepackage[caption=false,font=normalsize,labelfont=sf,textfont=sf]{subfig}
\usepackage{url} 
\usepackage{comment} 
\usepackage{bm}
\usepackage{multicol}
\usepackage{siunitx}
\usepackage{booktabs}
\usepackage{multirow}
\usepackage{bbm}
\usepackage{mathtools} 
\usepackage{etoolbox}
\usepackage{color}
\usepackage{authblk}
\usepackage{pifont}

\newcommand{\cmark}{\ding{51}}%
\newcommand{\xmark}{\ding{55}}%

\usepackage{tikz}

\newcommand{\RA}[1]{\textcolor{black}{#1}} 
\newcommand{\RB}[1]{\textcolor{black}{#1}} 
\newcommand{\RE}[1]{\textcolor{black}{#1}} 
\newcommand{\RC}[1]{\textcolor{black}{#1}} 
\newcommand{\RD}[1]{\textcolor{black}{#1}} 
\newcommand{\myvector}[1]{\boldsymbol{#1}}
\newcommand{\mytensor}[1]{\boldsymbol{\mathrm{#1}}}

\begin{document}

\title{Privacy-Preserving Image Classification Using Isotropic Network}

\author{AprilPyone MaungMaung and Hitoshi Kiya}
\affil{Tokyo Metropolitan University, 6--6 Hino Tokyo, Japan}

\maketitle

\begin{abstract}
In this paper, we propose a privacy-preserving image classification method that uses encrypted images and an isotropic network such as the vision transformer. The proposed method allows us not only to apply images without visual information to deep neural networks (DNNs) for both training and testing but also to maintain a high classification accuracy. In addition, compressible encrypted images, called encryption-then-compression (EtC) images, can be used for both training and testing without any adaptation network. Previously, to classify EtC images, an adaptation network was required before a classification network, so methods with an adaptation network have been only tested on small images. \RE{To the best of our knowledge, previous privacy-preserving image classification methods have never considered image compressibility and patch embedding-based isotropic networks.} In an experiment, the proposed privacy-preserving image classification was demonstrated to outperform state-of-the-art methods even when EtC images were used in terms of classification accuracy and robustness against various attacks under the use of two isotropic networks: vision transformer and ConvMixer.
\end{abstract}


\section{Introduction}

Deep learning has been deployed in many applications including security-critical ones such as biometric authentication and medical image analysis. Nevertheless, training a deep learning model requires a huge amount of data, efficient algorithms, and fast computing resources such as graphics processing units (GPUs) and tensor processing units (TPUs). Generally, data contains sensitive information and it is difficult to train a deep learning model while preserving privacy. For example, hospitals are not able to train a model on aggregated data due to regulations such as the Health Insurance Portability and Accountability Act (HIPAA) and General Data Protection Regulation (GDPR). In particular, data with sensitive information cannot be transferred to untrusted third-party cloud environments (cloud GPUs and TPUs) even though they provide a powerful computing environment. \RE{Therefore, it has been challenging to train/test deep learning models in cloud environments while preserving privacy.}

To address the privacy issue, researchers have proposed various solutions. Federated learning allows us to train a global model in a distributed manner in which only gradient updates are uploaded to a central server~\cite{kmyrsb16}. However, in many settings, data owners may not cooperate, and there is no aggregating server.

Cryptographic methods such as fully homomorphic encryption are still computationally expensive~\cite{shokri2015privacy}, and moreover, the encrypted images not only are uncompressible but also cannot be directly applied to models. For these reasons, various learnable perceptual encryption methods have been studied so far for various applications~\cite{kiya2022overview,2018-ICCETW-Tanaka,madono2020block,2019-Access-Warit,2020-Arxiv-Maung} that have been inspired by encryption methods for privacy-preserving photo cloud sharing services~\cite{2019-TIFS-Chuman}. \RB{Learnable image encryption is encryption that allows us not only to generate visually protected images to protect personally identifiable information included in an image such as an individual or the time and location of the taken photograph but to also apply encrypted images to a machine learning algorithm in the encrypted domain.}

Particularly, in the context of privacy-preserving image classification scenarios, learnable image encryption with an adaptation network~\cite{2018-ICCETW-Tanaka,madono2020block} and without an adaptation network~\cite{2019-Access-Warit} have been proposed. The adaptation network in~\cite{madono2020block} contains sub-networks based on the number of blocks in an image in addition to a classification network. Therefore, the number of parameters will be increased if the number of blocks in an image is increased. In the same line of research, other lightweight encoding schemes such as~\cite{ito2021image,huang2020instahide,yala2021neuracrypt} have also been put forward. Lightweight encoding schemes have security concerns as described in~\cite{carlini2021private,carlini2021neuracrypt}. Moreover, none of them consider compression efficiency even though compressed images are preferred for transmission.

\begin{figure}[t]
\centerline{\includegraphics[width=18.5pc]{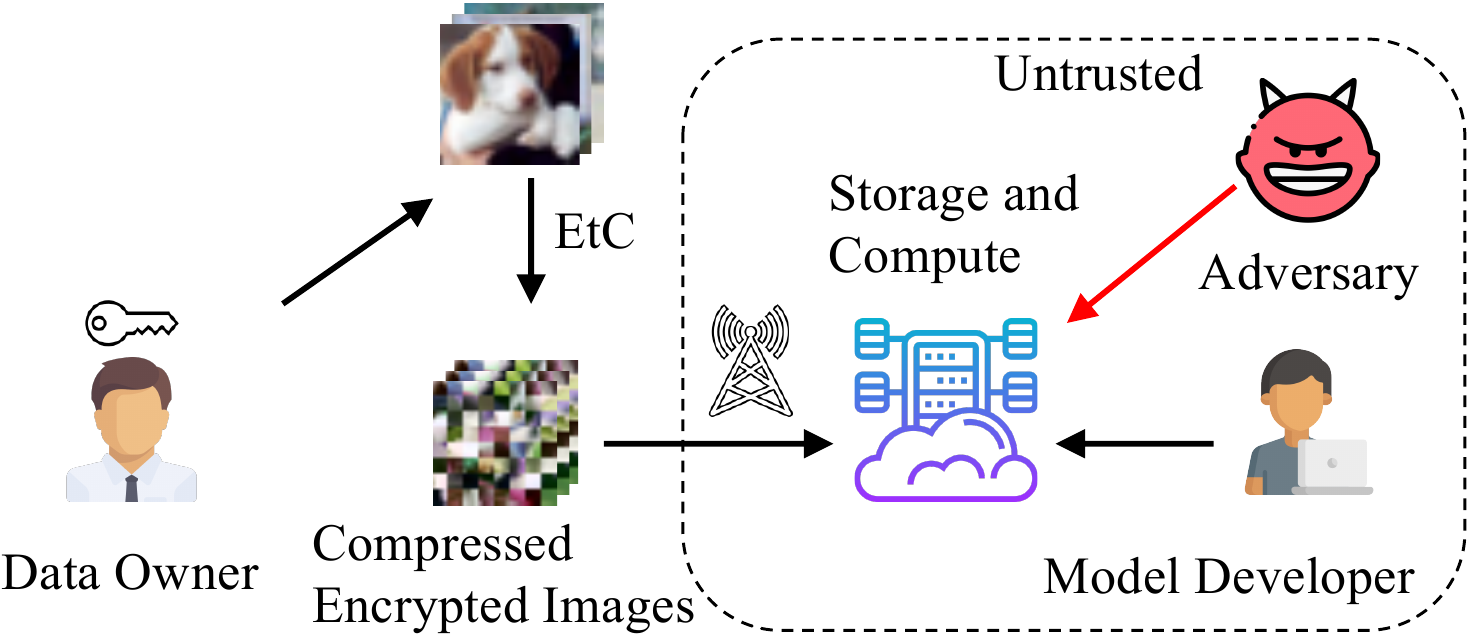}}
\caption{Scenario of classifying encrypted images.\label{fig:scenario}}
\end{figure}

We consider a scenario in Fig.~\ref{fig:scenario}, where a data owner encrypts and compresses images by applying encryption-then-compression (EtC) methods~\cite{2019-TIFS-Chuman}. Then, the compressed encrypted images are transmitted to an untrusted cloud provider for storage and computing, where encrypted images are used to train a model by a model developer. To improve the issues that the conventional methods have, we propose the combined use of encryption-then-compression (EtC) images~\cite{2019-TIFS-Chuman} and an isotropic network. Isotropic networks such as the vision transformer are well known to have a higher performance than conventional convolutional neural networks in some settings, and they include the operation of patch embedding. Patch embedding is expected to reduce the influence of image encryption on the classification accuracy because the patch embedding operation can express encryption steps as a transformation.

This paper and the method proposed in it have the following contributions:
\begin{itemize}
\item achieving the highest accuracy among previous privacy-preserving image classification methods with perceptual encryption,
\item outperforming previous privacy-preserving methods in terms of robustness against attacks and compressibility in addition to classification accuracy, and
\item conducting EtC image classification experiments with two isotropic networks, ConvMixer~\cite{anonymous2022patches} and vision transformer (ViT)~\cite{dosovitskiy2020image}, on CIFAR-10~\cite{2009-Report-Krizhevsky} and Imagenette, a subset of ImageNet~\cite{imagenette}.
\end{itemize}
\RE{To the best of our knowledge, previous privacy-preserving image classification methods have never considered image compressibility and patch embedding-based isotropic networks.}

\section{Related Work}
\subsection{Privacy-Preserving Image Classification}
Generally, privacy-preserving machine learning addresses three issues: the (1) privacy of datasets, (2) privacy of models, and (3) privacy of models' outputs~\cite{shokri2015privacy}. For image classification tasks, we focus on the privacy of datasets in this paper. There are two approaches to preserving the privacy of datasets: learnable encryption with a secret key and image encoding without a secret key. Image encoding approaches do not consider the ability to decode in contrast to image encryption methods.

Learnable encryption encrypts images with a secret key so that visual information in encrypted images is not perceptible to humans while maintaining the ability to classify encrypted images with a model. Tanaka first introduced block-wise learnable image encryption (LE) with an adaptation layer that is used prior to the classifier to reduce the influence of image encryption~\cite{2018-ICCETW-Tanaka}. Another encryption method is pixel-wise encryption (PE) in which negative-positive transformation and color component shuffling are applied without using any adaptation layer~\cite{2019-Access-Warit}. However, both block-wise~\cite{2018-ICCETW-Tanaka} and pixel-wise~\cite{2019-Access-Warit} encryption methods can be attacked by ciphertext-only attacks~\cite{chang2020attacks}. To enhance the security of encryption, Tanaka's method was extended by adding a block scrambling step and utilizing different block keys for the pixel encryption operation~\cite{madono2020block} (hereinafter denoted as ELE). The adaptation network of the ELE method was also applied to EtC images with a small block size. However, the classification accuracy was low, and the EtC images were not JPEG-compliant.

Image encoding approaches for privacy preserving encode images to hide visual information.
One method employs a transformation network with U-Net that is also trained by using a pre-trained classifier to protect visual information during inference time~\cite{ito2021image}. However, this image encoding method has never been tested on larger images (e.g., with a dimension of $3 \times 224 \times 224$). Recently, another encoding method, InstaHide, encodes images by mixing them with other images and applying a pixel-wise sign-flipping mask~\cite{huang2020instahide}. However, it was demonstrated that InstaHide-encoded images can be reconstructed~\cite{carlini2021private}.
Similarly, NeuraCrypt encodes images by using a random neural network with positional encoding~\cite{yala2021neuracrypt}. However, an attack on NeuraCrypt was also released to match encoded images and plain ones completely (NeuraCrypt Challenge 1)~\cite{carlini2021neuracrypt}. All in all, these encoding methods do not consider compression efficiency even though a large volume of images is practically transmitted in a compressed form.

In this paper, we focus on classifying EtC images by using an isotropic network. Although EtC images were previously classified with an adaptation network~\cite{madono2020block}, there are a number of issues: low accuracy, \RA{vulnerability to attacks, the need for an adaptation network, a lack of compressibility}, and unknown applicability to a large image size. We aim to address these issues in this paper. Table~\ref{tab:encryption} summarizes the previous related work and the proposed classification.

\begin{table}
 \centering
 \caption{Summary of privacy-preserving image classification with different encryption methods and \RA{proposed one}. (\cmark) denotes ``Yes,'' and (\xmark) denotes ``No.''\label{tab:encryption}}
\resizebox{\columnwidth}{!}{%
 \begin{tabular}{cccccc}
 \toprule
 {Requirements} & LE~\cite{2018-ICCETW-Tanaka} & ELE~\cite{madono2020block} & PE~\cite{2019-Access-Warit} & EtC~\cite{madono2020block} & Proposed\\
 \midrule
 High Accuracy & {\cmark} & {\xmark} & {\cmark} & {\xmark} & {\cmark}\\[0.5em]
 Robustness & {\xmark} & {\cmark} & {\xmark} & {\cmark} & {\cmark}\\[0.5em]
 Without Adaptation & {\xmark} & {\xmark} & {\cmark} & {\xmark} & {\cmark}\\[0.5em]
 \RA{Compressibility} & {\xmark} & {\xmark} & {\xmark} & {\xmark} & {\cmark}\\[0.5em]
 Large Input Size$^{\dagger}$ & {\xmark} & {\xmark} & {\xmark} & {\xmark} & {\cmark}\\[0.5em]
 \bottomrule
 \multicolumn{6}{l}{$^{\dagger}$ Large images were never tested on conventional methods.}
 \end{tabular}
 }
\end{table}

\subsection{EtC Images}
In this paper, we use EtC images as compressible encrypted images which were proposed for privacy-preserving cloud photo sharing services~\cite{2019-TIFS-Chuman}. EtC images have not only almost the same compression performance as plain images but also robustness against various ciphertext-only attacks~\cite{2019-TIFS-Chuman}. Here, we summarize a method for generating EtC images that was proposed in~\cite{2019-TIFS-Chuman}.

\begin{figure}[t]
\centerline{\includegraphics[width=18.5pc]{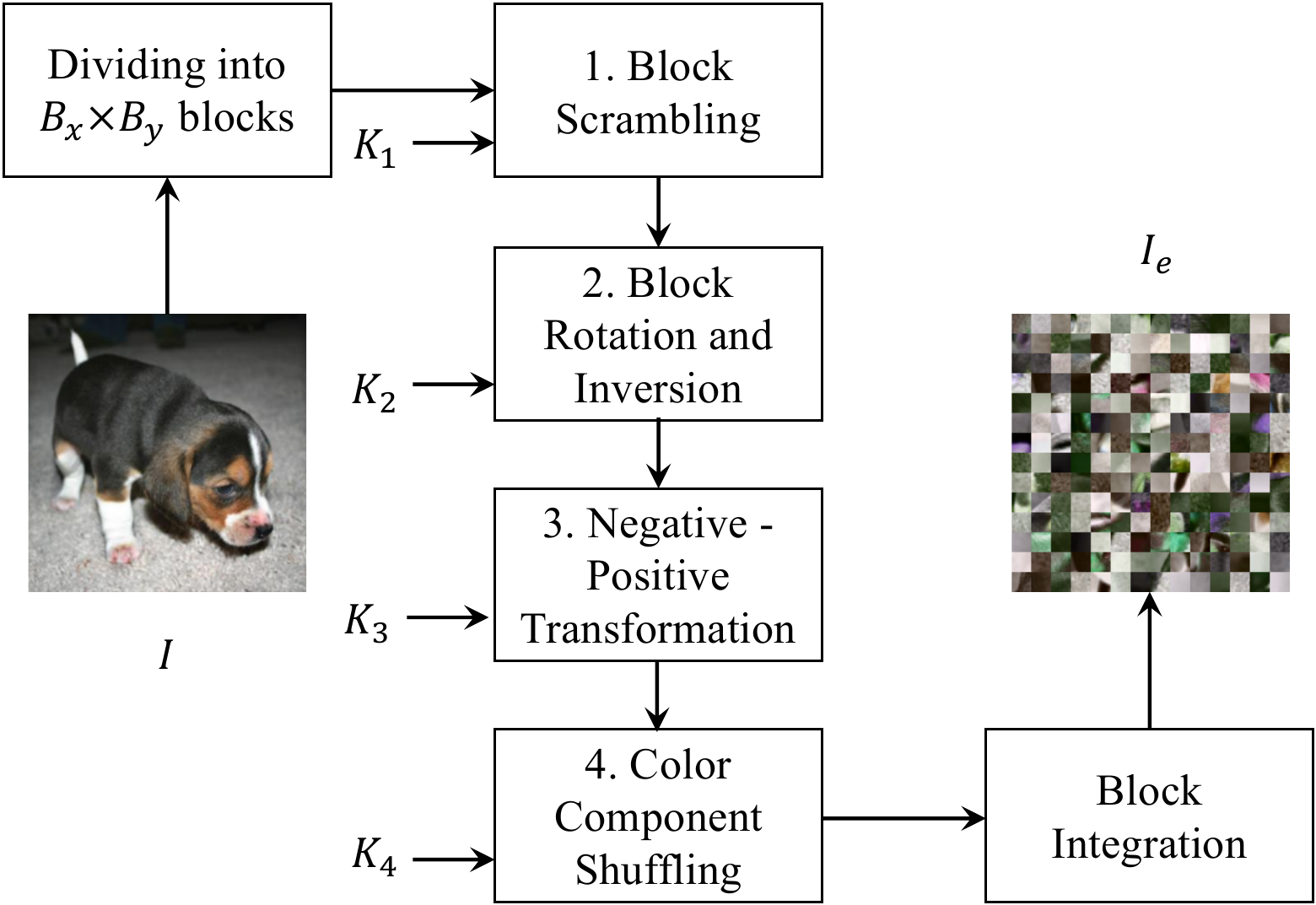}}
\caption{Procedure of EtC image generation~\cite{2019-TIFS-Chuman}.\label{fig:etc}}
\end{figure}

Figure~\ref{fig:etc} depicts the procedure of \RB{block-wise encryption}. A three-channel (RGB) color image ($I$) with $X \times Y$ pixels is divided into non-overlapping blocks each with $B_x \times B_y$. Then, four encryption steps are carried out on the divided blocks as follows.
\begin{enumerate}
 \item Randomly permute the divided blocks by using a random integer generated by a secret key $K_1$.
 \item Rotate and invert each block randomly by using a random integer generated by a key $K_2$.
 \item Apply negative-positive transformation to each block by using a random binary integer generated by a key $K_3$, where $K_3$ is commonly used for all color components. A transformed pixel value in the $i$\textsuperscript{th} block, $p'$, is calculated using
 \begin{equation}
 p' = \left\{
 \begin{array}{ll}
 p & (r(i) = 0)\\
 p \oplus (2^L - 1) & (r(i) = 1),
 \end{array}
 \right.
 \end{equation}
 where $r(i)$ is a random binary integer generated by $K_3$, $p$ is the pixel value of the original image with $L$ bits per pixel ($L=8$ is used in this paper), and $\oplus$ is the bitwise exclusive-or operation. The value of the occurrence probability $\mathrm{P}(r(i)) = 0.5$ is used to invert bits randomly.
 \item Shuffle three color components in each block by using an integer randomly selected from six integers generated by a key $K_4$.
\end{enumerate}

Then, integrate the encrypted blocks to form an encrypted image $I_e$. Note that block size $B_x = B_y = 16$ is enforced to be JPEG compatible.

\subsection{Isotropic Networks}
Following the success of the vision transformer (ViT), a number of isotropic networks (with the same depth and resolution across different layers in the network) have been proposed such as MLP-Mixer, ResMLP, CycleMLP, gMLP, vision permutator, and ConvMixer.
\RD{For experiments, we chose ViT~\cite{dosovitskiy2020image} and ConvMixer~\cite{anonymous2022patches} as two examples of isotropic networks for classifying EtC images.
Other isotropic networks that have patch embedding are also expected to provide a similar performance.
}

\begin{figure*}[t]
\centering
\subfloat[]{\includegraphics[width=18.5pc]{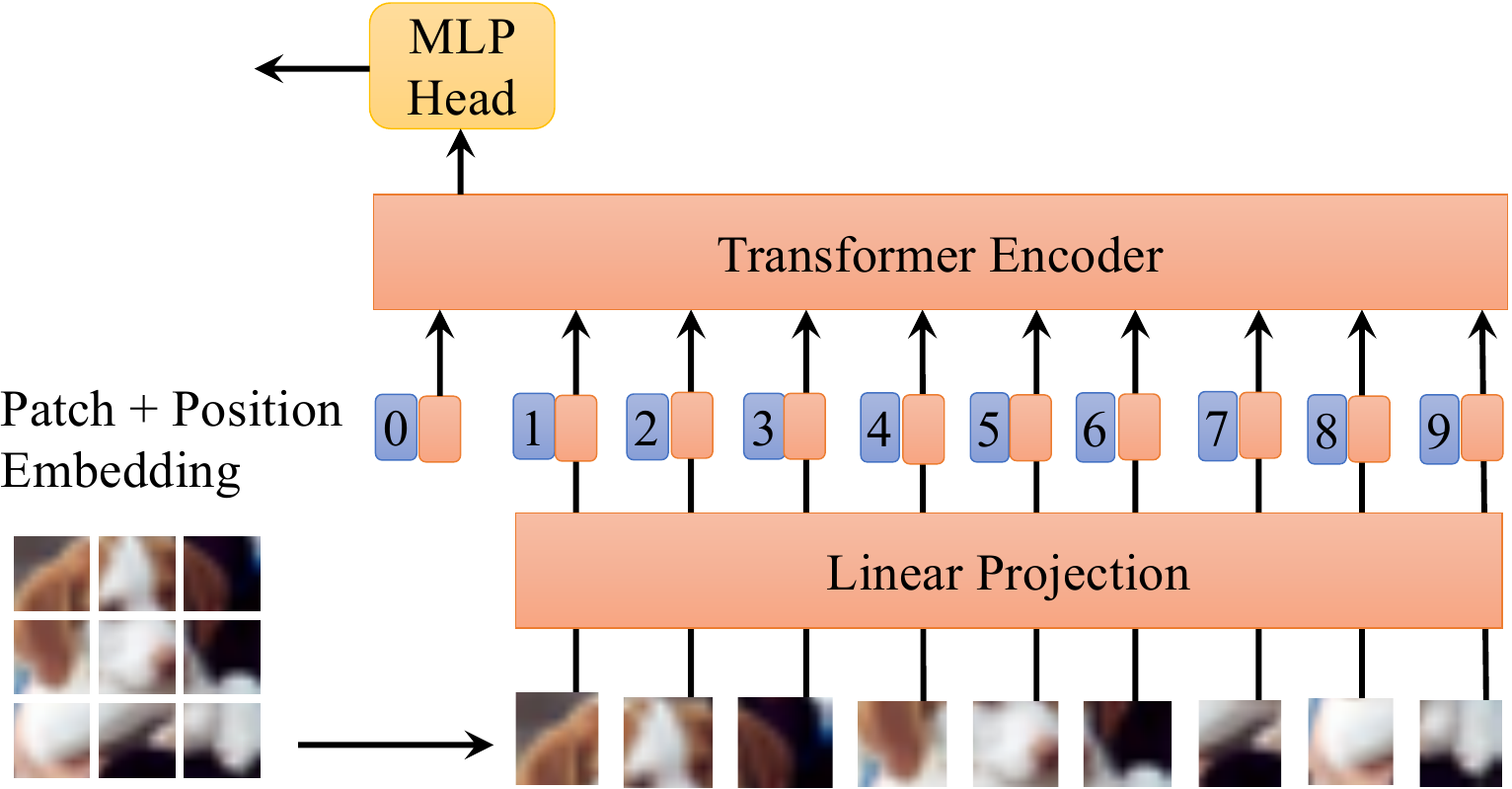}%
\label{fig:vit}}
\hfil
\subfloat[]{\includegraphics[width=36.5pc]{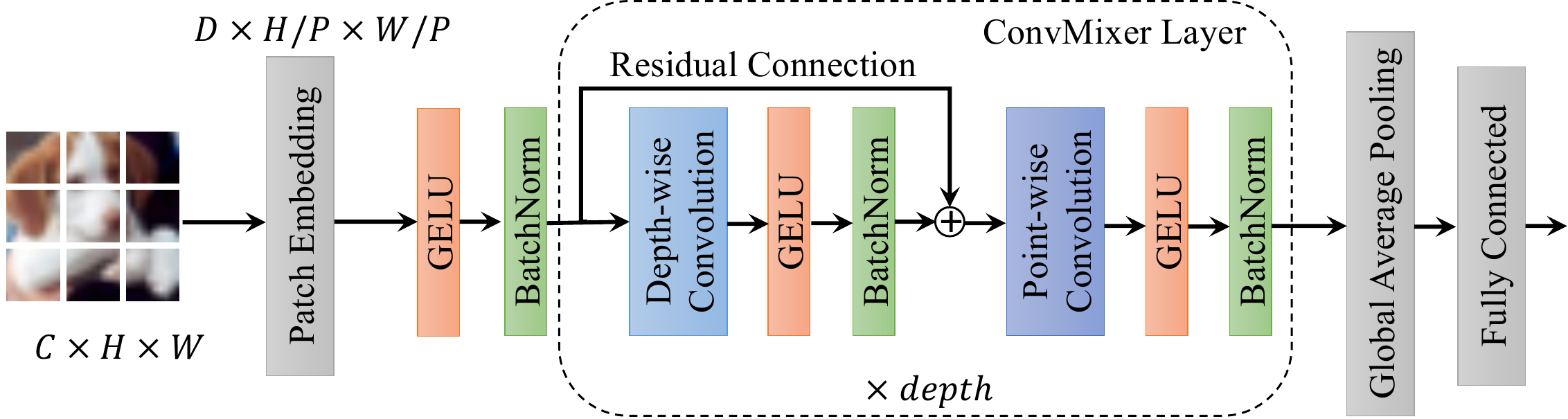}%
\label{fig:convmixer}}
\caption{Two isotropic networks. (a) Architecture of ViT~\cite{dosovitskiy2020image}. (b) Architecture of ConvMixer~\cite{anonymous2022patches}. In (a), Patch embedding and position embedding are added together, and resulting embedding is used as input to transformer encoder. In (b), Image (with dimension of $C \times H \times W$) is transformed by patch embedding to output $D \times H/P \times W/P$ ($D$ is embedding dimension, and $P$ is patch size) with non-linear activation (GELU) and batch normalization (BatchNorm) followed by multiple ConvMixer layers ($\times depth$) before global average pooling and fully-connected layer for classification.\label{fig:isotropic}}
\end{figure*}

ViT that utilizes patch embedding and position embedding (see Fig.~\ref{fig:vit}), and ConvMixer~\cite{anonymous2022patches} does not have the patch-order invariance because of convolutions along with patches (see Fig.~\ref{fig:convmixer}).
With the use of a large kernel size (receptive field), ConvMixer is also able to achieve a competitive classification performance.
In a later section, we will discuss the theoretical analysis on why EtC images can be classified by isotropic networks.

\section{Proposed Image Classification Using Isotropic Network}
A novel privacy-preserving image classification method using an isotropic network is proposed here.
\subsection{Requirements}
For privacy-preserving image classification, we aim to achieve three requirements:
\begin{itemize}
 \item High Classification Accuracy: A model trained by encrypted images should be approximately as good as a model trained by using plain images.
 \item Robustness Against Attacks: Visual information of plain images should not be restored from encrypted images.
 \item Compressibility: Images are transmitted in a compressed form (e.g., JPEG), and therefore, compression efficiency should be maintained even when using encrypted images.
\end{itemize}

\subsection{Overview}
\begin{figure}[t]
\centerline{\includegraphics[width=18.5pc]{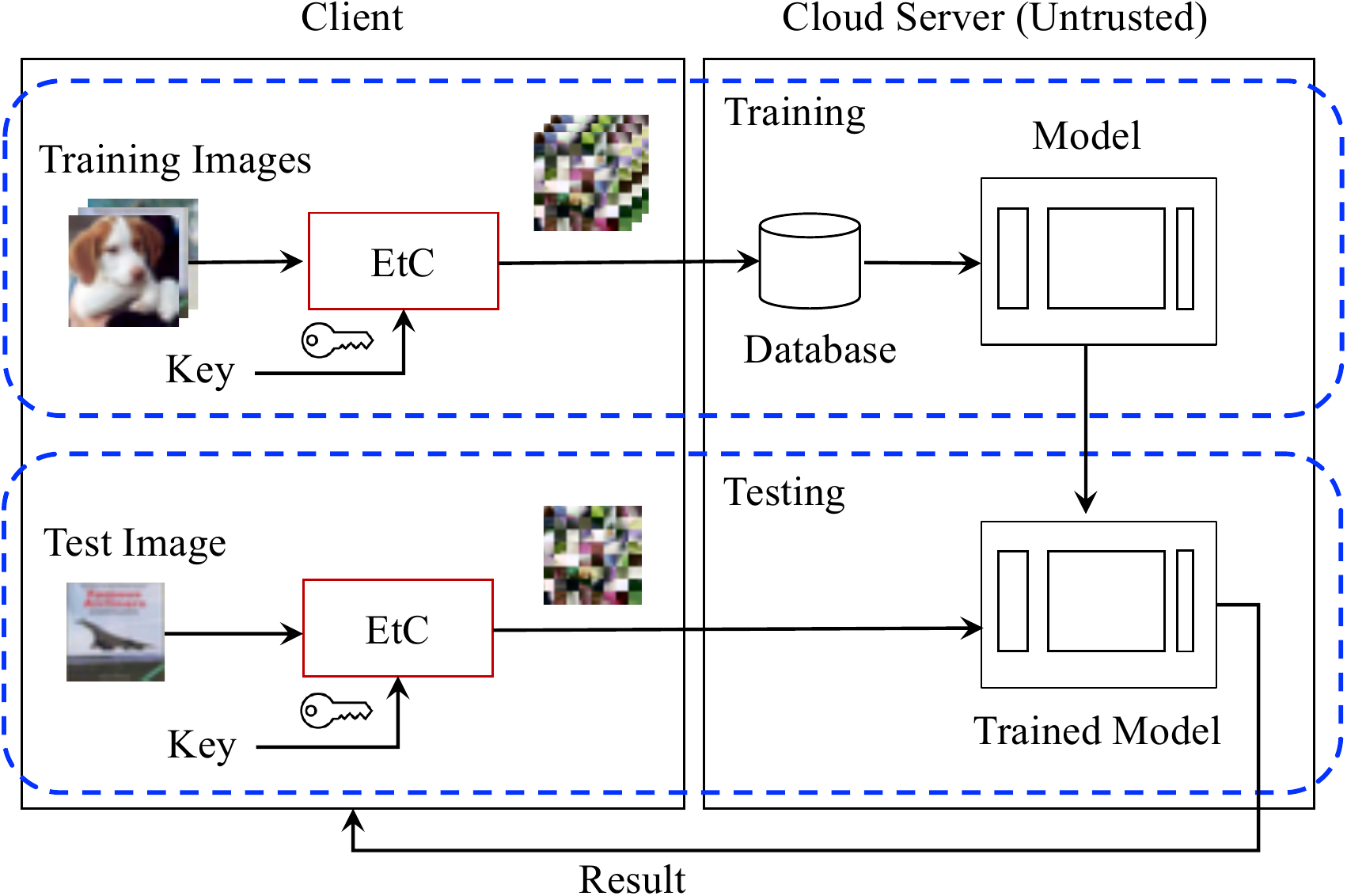}}
\caption{Overview of proposed privacy-preserving image classification.\label{fig:overview}}
\end{figure}

To satisfy the above requirements, we propose using EtC images~\cite{2019-TIFS-Chuman} and an isotropic network. An overview of the proposed privacy-preserving image classification scenario is illustrated in Fig.~\ref{fig:overview}.

A user (client) encrypts training images to generate EtC images with a secret key and sends the encrypted images to a cloud provider (untrusted) for storage and training a model. Then, a model is trained by using the encrypted images without any perceptible visual information. For inference (testing), the same secret key is required to encrypt test images, and the trained model classifies the encrypted test images. Therefore, in both training and testing a model, the cloud provider cannot know the visual information of images. In addition, the use of EtC images allows us to compress encrypted images for storage and transmission. \RA{To the best of our knowledge, isotropic networks have never been used for privacy-preserving image classification before.}

\subsection{Threat Models}
As we focus on the privacy of datasets in an image classification scenario, the goal of an adversary is to recover visual information on encrypted images. Encrypted images are transferred to an untrusted cloud provider for storage and training/testing a model as in Figs.~\ref{fig:scenario} and~\ref{fig:overview}. Therefore, we assume the adversary has access to encrypted images and knows the encryption algorithm but not the secret key. \RB{In other words, we assume that the adversary can carry out a ciphertext-only attack (COA) only from encrypted images.} In addition, we also assume the adversary knows the distribution of the dataset; thus, the adversary may prepare synthetic examples in conducting an attack.

\subsection{Training}
We use different training strategies for ViT and ConvMixer. As ViT provides a high classification accuracy when pre-training on a large amount of data, we take advantage of a ViT pre-trained model with the Imagenet21k dataset. In contrast, we train a ConvMixer without any pre-training.

ViT employs positional embedding to learn the relative positions of patches, which motivates us to use EtC images.
Note that the patch size in ViT and the block size in EtC have to be the same to maintain a high accuracy.

In contrast, ConvMixer does not use any positional embedding. However, the use of a large kernel size allows us to learn EtC images with a ConvMixer even when the patch size and the block size are different.

\subsection{Testing}
The same secret key as the key used for training a model is applied for encrypting test images before sending them to an untrusted cloud provider as shown in Fig.~\ref{fig:overview}.
The model at the provider classifies encrypted images without being aware of any visual information.

\subsection{Analysis of Isotropic Networks}
\RD{Isotropic networks such as ViT~\cite{dosovitskiy2020image} and ConvMixer~\cite{anonymous2022patches} have a patch embedding structure. In this paper, we point out for the first time that patch embedding enables us to reduce the influence of encryption.
}

\RD{
For example, in ViT~\cite{dosovitskiy2020image}, an image tensor $\mytensor{x} \in [0, 1]^{C \times H \times W}$ is reshaped into a sequence of flattened patches $\myvector{x}_{p} = \{\myvector{x}_{p}^{1}, \myvector{x}_{p}^{2}, \ldots, \myvector{x}_{p}^{i}, \ldots, \myvector{x}_{p}^{N}\} \in [0, 1]^{N \times (P^{2} \cdot C)}$, where $C$ is the number of channels, $H$ is the height of an image, $W$ is the width of the image, $P$ is the patch size, and $N$ is the number of patches ($N = HW/P^2$). For simplicity, we assume that a patch is a square shape $(P, P)$, and $H$ and $W$ are divisible by $P$. Then, ViT~\cite{dosovitskiy2020image} uses patch embedding and positional embedding (see Fig.~\ref{fig:vit}). In patch embedding, flattened patches are mapped to vectors with dimensions of $D$, and in position embedding, the position information $\myvector{e}^{i}_{\text{pos}}$ is embedded into a patch as
\begin{equation}
 \begin{array}{l}
 \myvector{z}^{i}_{0} = \myvector{x}^{i}_{p}\mytensor{E} + \myvector{e}^{i}_{\text{pos}},\\
 \mytensor{E} \in \mathbb{R}^{(P^2 \cdot C) \times D}, \myvector{e}^{i}_{\text{pos}} \in \mathbb{R}^{D}.
\end{array}
\end{equation}
}

\RD{
Let us consider a matrix $\mytensor{E}_{\text{pos}}$ consisting of position information $\myvector{e}^{i}_{\text{pos}}$ used in patch embedding as
\begin{equation}
 \mytensor{E}_{\text{pos}} = \begin{pmatrix}\begin{pmatrix}\myvector{e}^{1}_{\text{pos}}\end{pmatrix}^{\mathrm{T}} & 
 \begin{pmatrix}\myvector{e}^{2}_{\text{pos}}\end{pmatrix}^{\mathrm{T}} & \ldots & \begin{pmatrix}\myvector{e}^{N}_{\text{pos}}\end{pmatrix}^{\mathrm{T}}
\end{pmatrix}^{\mathrm{T}}.
\end{equation}
The permutation of rows in $\mytensor{E}_{\text{pos}}$ corresponds to the block scrambling (step $1$ in Fig.~\ref{fig:etc}) in image encryption. Therefore, a positional embedding matrix learned from block-scrambled images can ideally be expressed as
\begin{equation}
 \mytensor{E'}_{\text{pos}} = \mytensor{E}_{1}\mytensor{E}_{\text{pos}},
\end{equation}
where $\mytensor{E}_{1}$ is a permutation matrix decided by using key $K_1$ used for the block scrambling.
}

\RD{
If every patch (block) is transformed with the same key, encryption step 2 (block rotation and inversion), step 3 (negative-positive transformation under normalization), and step 4 (color component shuffling) can also be expressed as a pixel shuffling operation with sign flipping. Thus, a patch embedding parameter $\mytensor{E'}$ learned from encrypted images is ideally given as
\begin{equation}
 \mytensor{E'} = \mytensor{E}_{2}\mytensor{E},
\end{equation}
where $\mytensor{E}_2$ is a matrix decided by encryption operations used for steps 2, 3, and 4.
}

\RD{
Step 3 corresponds to sign flipping under the use of normalization. A transformed pixel value $p'$ at $[0, 1]$ scale can also be expressed as
 \begin{equation}
 p' = \left\{
 \begin{array}{ll}
 p & (r(i) = 0)\\
 1 - p & (r(i) = 1).
 \end{array}
 \right. \label{eqn:np}
 \end{equation}
In the normalization used in this paper, when $r(i) = 1$ in Eq.~\eqref{eqn:np}, a pixel $p$ is replaced with $p'$ as
\begin{align}
 p' &= \frac{p - 0.5}{0.5} \nonumber \\
 &= \frac{(1 - p) - 0.5}{0.5} \nonumber \\
 &= \frac{0.5 - p}{0.5} \nonumber \\
 &= - \left(\frac{p - 0.5}{0.5}\right) \nonumber \\
 &= - p'.
\end{align}
 }

\RD{
Note that each patch is not transformed with the same key for encryption steps $2$, $3$, and $4$ in the EtC scheme. Therefore, the above transformation in $\mytensor{E}$ does not exactly reflect the EtC scheme, but there is a high compatibility between the patch embedding and the encryption. Accordingly, the patch embedding in isotropic networks is expected to reduce a part of the influence of encryption as well as the use of an adaptation network, and the use of an isotropic network provides a high classification accuracy. Other isotropic networks with patch embedding such as MLP-Mixer and ResMLP are also expected to provide a high classification accuracy.
}

\subsection{Robustness Evaluation}
\RD{
In existing learnable encryption methods, robustness against various attacks, which aim to restore visual information on plain images, has been evaluated under cipher-text attacks, so the proposed method is compared with conventional methods in terms of robustness against cipher-text attacks in this paper.
}

\RD{Robustness of EtC images against not only brute-force attack but also jigsaw puzzle solver attacks as ciphertext-only attacks were evaluated in~\cite{2019-TIFS-Chuman}. An EtC image has almost the same correlation among pixels in each block as that of the original image, whose property enables to efficiently compress images. Therefore, an attacker can utilize the correlation to decrypt the image in some way, so security of the encryption against jigsaw puzzle solver attacks was discussed in~\cite{2019-TIFS-Chuman} in addition to brute-force attack.}

\RD{In contrast, recently, novel attack methods for restoring visual information have been proposed that use deep neural networks such as in~\cite{madono2021gan,ito2021image}.
Accordingly, we consider the feature reconstruction attack (FR-Attack)~\cite{chang2020attacks} that exploits the local properties of an encrypted image to reconstruct visual information from encrypted images. Furthermore, with a synthetic dataset and encrypted images, the adversary may carry out a GAN-based attack (GAN-Attack)~\cite{madono2021gan}. As the distribution of the dataset is known, we also consider that the adversary may prepare exact pairs of plain images and encrypted ones with different multiple keys to learn a transformation model, i.e., the inverse transformation network attack (ITN-Attack)~\cite{ito2021image}.
}

\robustify\bfseries
\sisetup{table-parse-only,detect-weight=true,detect-inline-weight=text,round-mode=places,round-precision=2}
\begin{table*}
\centering
\caption{\RB{Classification accuracy (\SI{}{\percent}) and robustness of models trained by using encrypted images or plain ones compared with state-of-the-art methods. (N/A) denotes ``not applicable.'' ($\dagger$) denotes that model was applied only to input dimension ($3 \times 32 \times 32$), ($\ddagger$) indicates that model was applied only to input dimension ($3 \times 224 \times 224$), and ($\ast$) denotes that model was not available for Imagenette dataset. Best results are in bold.\label{tab:results}}}
\begin{tabular}{ccccSSc}

 \toprule
 \multirow{2}{*}{Encryption} & \multirow{2}{*}{Model} & \multirow{2}{*}{\# Parameters} & \multirow{2}{*}{Block Size} & \multicolumn{2}{c}{Accuracy} & {Robustness}\\
 \cmidrule{5-6}
 & & {$\approx$($\times 10^6$)} & {($B_x = B_y$)} & {CIFAR-10} & {Imagenette}\\
 \midrule
 Plain & ViT-B\_16 & 85.81 & {N/A} & 99.06 & 99.64 & {No}\\
 EtC (Proposed) & ViT-B\_16 & 85.81 & 16 & 87.89 & \bfseries \num{90.62} & {Yes}\\ 
 Plain & ConvMixer-256/8 & 0.71 &{N/A} & 96.07 & {$\dagger$} & {No}\\
 EtC (Proposed) & ConvMixer-256/8 & 0.71 & 16 & \bfseries \num{92.72} & {$\dagger$} & {Yes}\\
 Plain & ConvMixer-384/8 & 1.47 &{N/A} & {$\ddagger$} & 96.10 & {No}\\
 EtC (Proposed) & ConvMixer-384/8 & 1.47 & 16 & {$\ddagger$} & 90.11 & {Yes}\\
 \midrule
 Plain & ShakeDrop & 28.48 & {N/A} & 96.70 & {$\ast$} & {No}\\
 LE~\cite{2018-ICCETW-Tanaka} & Adaptation + ShakeDrop & 62.97 & {4} & 94.49 & {$\ast$} & {No}\\
 ELE~\cite{madono2020block} & Adaptation + ShakeDrop & 62.97 & {4} & 83.06 & {$\ast$} & {Yes}\\
 EtC~\cite{madono2020block} & Adaptation + ShakeDrop & 62.97 & {4} & 89.09 & {$\ast$} & {Yes}\\
 Plain & ResNet-18 & 11.17 & {N/A} & 95.53 & {$\ast$} & {No}\\
 PE~\cite{2019-Access-Warit} & ResNet-18 & 11.17 & {N/A} & 91.33 & {$\ast$} & {No}\\
 \bottomrule
\end{tabular}
\end{table*}

\begin{figure*}[t]
\centering
\subfloat[]{\includegraphics[width=0.25\linewidth]{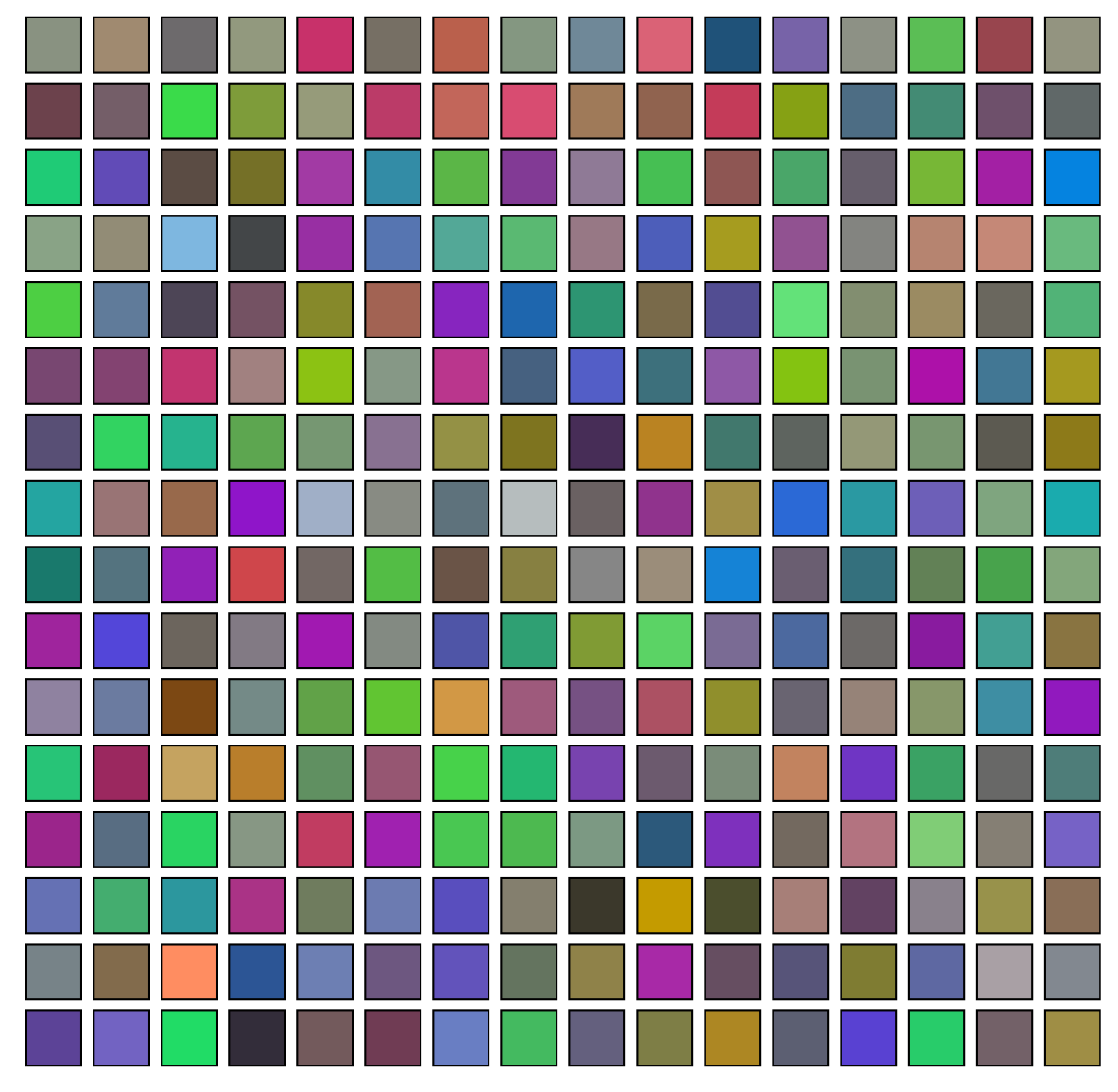}%
\label{fig:plain-weight}}
\hfil
\subfloat[]{\includegraphics[width=0.25\linewidth]{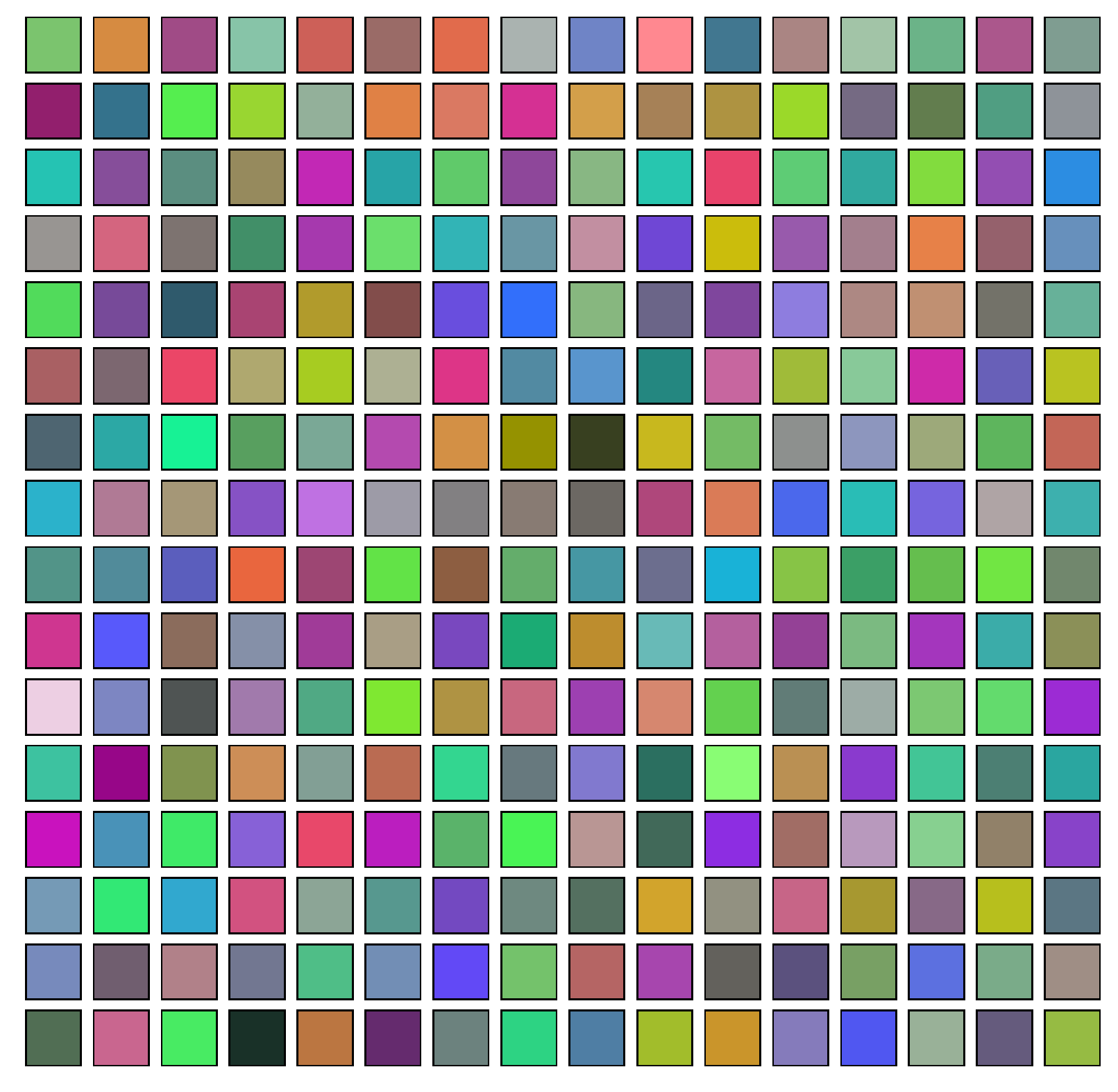}%
\label{fig:etc-weight}}
\hfil
\subfloat[]{\includegraphics[width=0.25\linewidth]{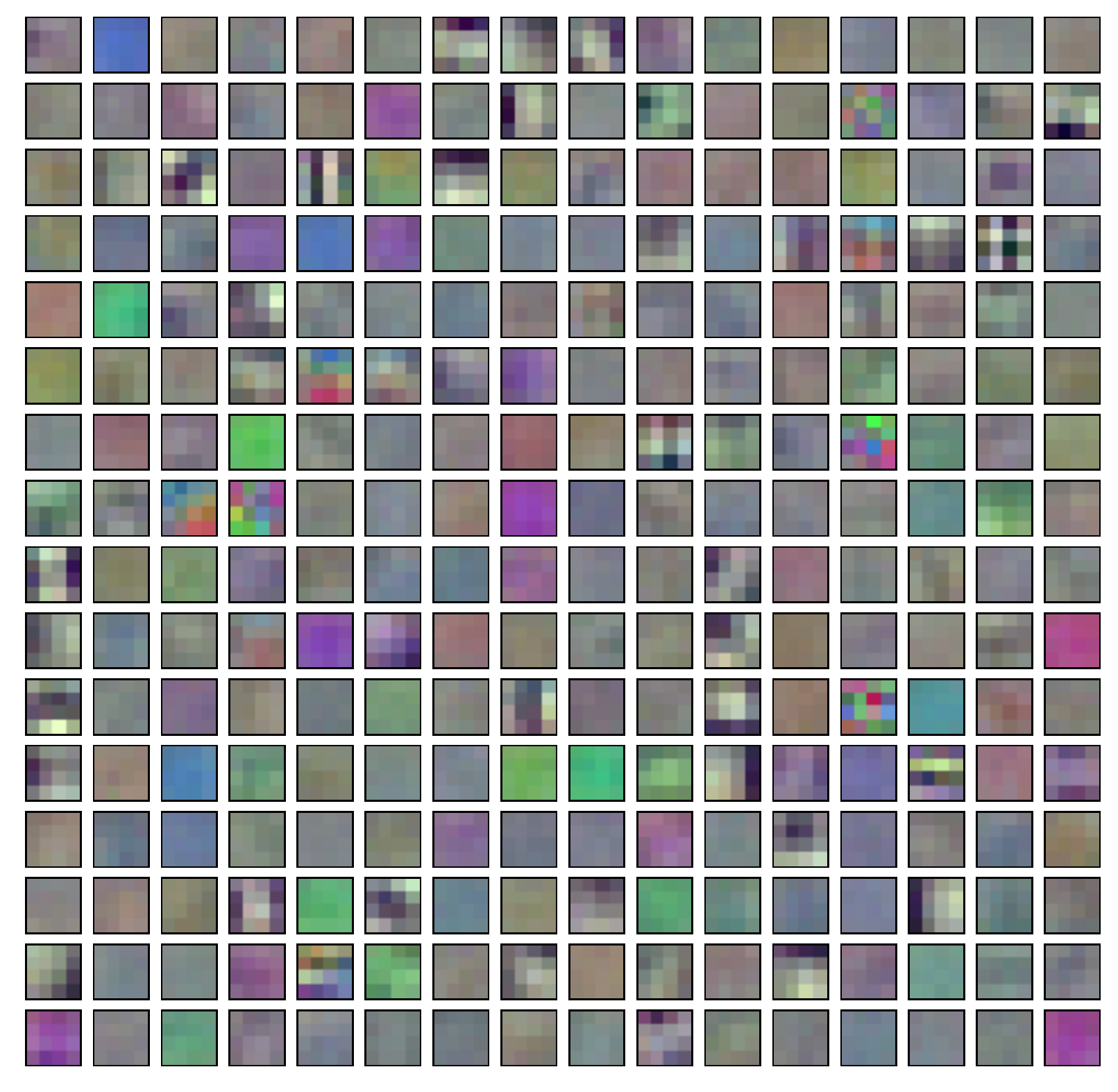}%
\label{fig:plain2-weight}}
\hfil
\subfloat[]{\includegraphics[width=0.25\linewidth]{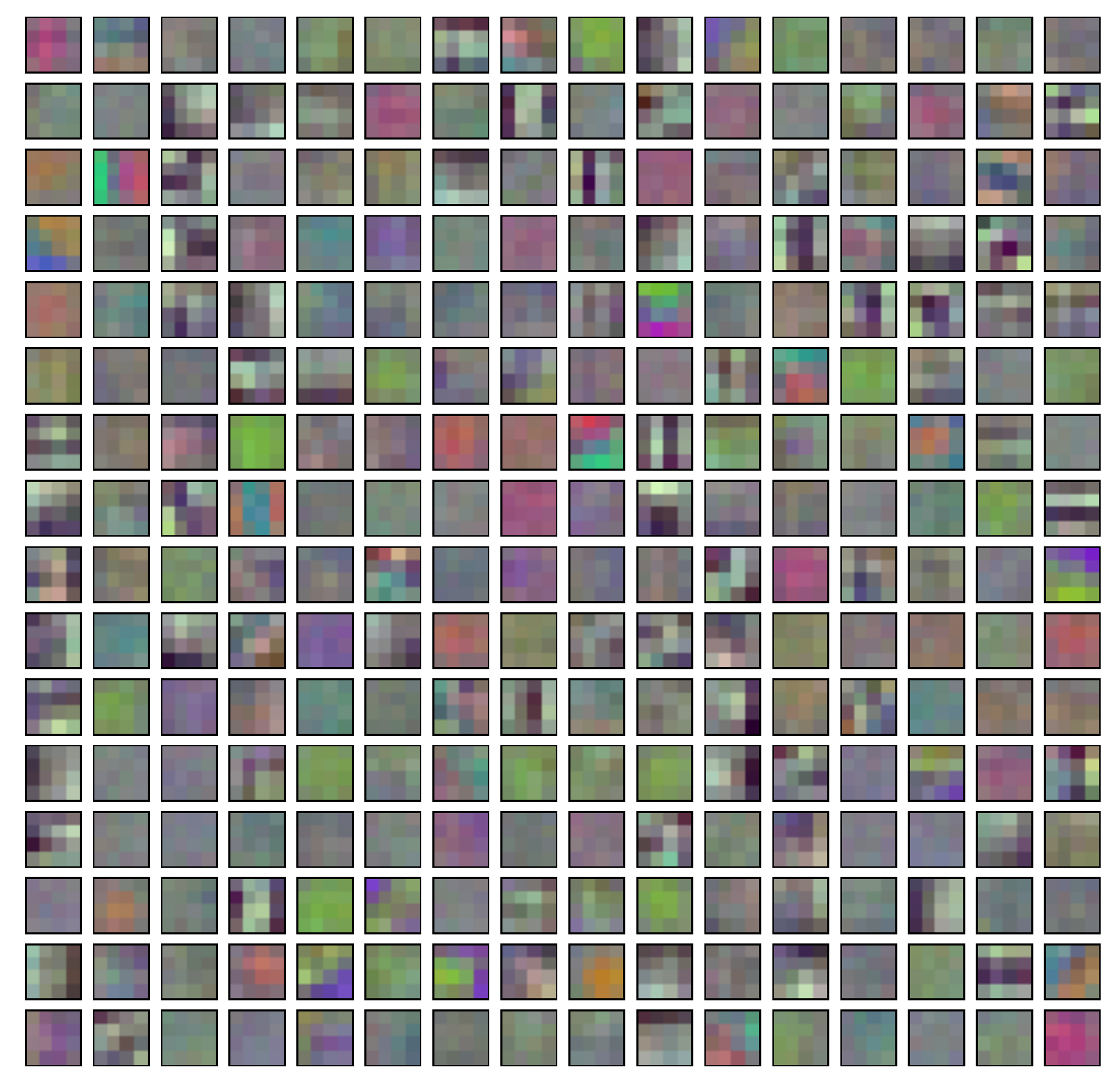}%
\label{fig:etc2-weight}}
\caption{\RA{Visualization of patch embedding weights of ConvMixer-256/8 for CIFAR-10 and ConvMixer-384/8 for Imagenette. (a) Embedding weights of plain images (ConvMixer-256/8). (b) Embedding weights of encrypted images (ConvMixer-256/8). (c) Embedding weights of plain images (ConvMixer-384/8). (d) Embedding weights of encrypted images (ConvMixer-384/8). ConvMixer-256/8 used patch size value of 1, and ConvMixer-384/8 used patch size value of 4.\label{fig:weight-visualize}}}
\end{figure*}

\begin{figure*}[t]
 \centering
 \begin{tabular}{ccccc}
 & LE~\cite{2018-ICCETW-Tanaka} & ELE~\cite{madono2020block} & PE~\cite{2019-Access-Warit} & EtC~\cite{2019-TIFS-Chuman}\\
 Encrypted &
 \begin{minipage}{2.0cm}
 \centering
 \includegraphics[width=2.0cm]{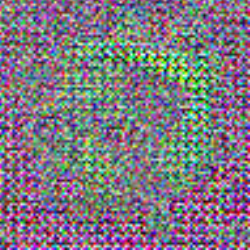}
 \end{minipage} &
 \begin{minipage}{2.0cm}
 \centering
 \includegraphics[width=2.0cm]{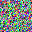}
 \end{minipage} &
 \begin{minipage}{2.0cm}
 \centering
 \includegraphics[width=2.0cm]{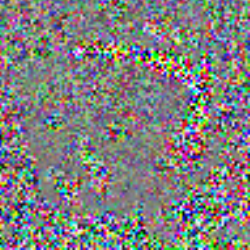}
 \end{minipage} &
 \begin{minipage}{2.0cm}
 \centering
 \includegraphics[width=2.0cm]{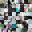}
 \end{minipage}\\
 & 0.037 & 0.002 & 0.041 & 0.047\\[5pt]
 FR-Attack~\cite{chang2020attacks} &
 \begin{minipage}{2.0cm}
 \centering
 \includegraphics[width=2.0cm]{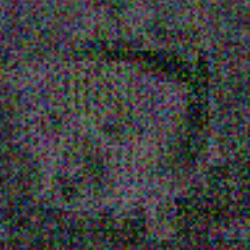}
 \end{minipage} &
 \begin{minipage}{2.0cm}
 \centering
 \includegraphics[width=2.0cm]{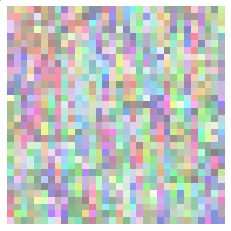}
 \end{minipage} &
 \begin{minipage}{2.0cm}
 \centering
 \includegraphics[width=2.0cm]{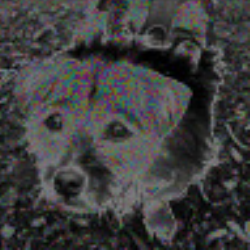}
 \end{minipage} &
 \begin{minipage}{2.0cm}
 \centering
 \includegraphics[width=2.0cm]{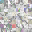}
 \end{minipage}\\
 & 0.101 & 0.055 & 0.303 & 0.042\\[5pt]
 GAN-Attack~\cite{madono2021gan} &
 \begin{minipage}{2.0cm}
 \centering
 \includegraphics[width=2.0cm]{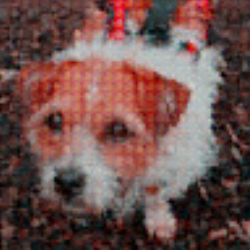}
 \end{minipage} &
 \begin{minipage}{2.0cm}
 \centering
 \includegraphics[width=2.0cm]{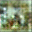}
 \end{minipage} &
 \begin{minipage}{2.0cm}
 \centering
 \includegraphics[width=2.0cm]{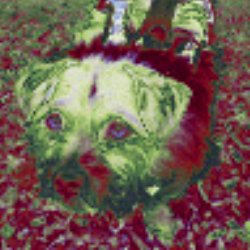}
 \end{minipage} &
 \begin{minipage}{2.0cm}
 \centering
 \includegraphics[width=2.0cm]{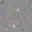}
 \end{minipage} \\
 & 0.864 & 0.009 & 0.109 & 0.018\\[5pt]
 ITN-Attack~\cite{ito2021image} &
 \begin{minipage}{2.0cm}
 \centering
 \includegraphics[width=2.0cm]{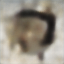}
 \end{minipage} &
 \begin{minipage}{2.0cm}
 \centering
 \includegraphics[width=2.0cm]{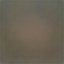}
 \end{minipage} &
 \begin{minipage}{2.0cm}
 \centering
 \includegraphics[width=2.0cm]{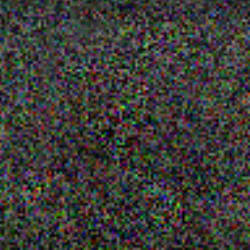}
 \end{minipage} &
 \begin{minipage}{2.0cm}
 \centering
 \includegraphics[width=2.0cm]{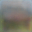}
 \end{minipage}\\
 & 0.281 & 0.071 & 0.017 & 0.025\\
 Plain & \multicolumn{4}{c}{
 \begin{minipage}{2.0cm}
 \centering
 \includegraphics[width=2.0cm]{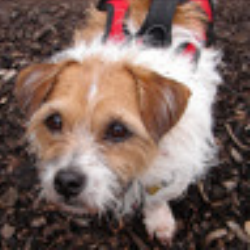}
 \end{minipage}
 }
 \end{tabular}
 \caption{Examples of restored images from encrypted ones. Partial results are taken from~\cite{ito2021image}. SSIM values are given under images.}
 \label{fig:sec-eval}
\end{figure*}

\section{Experiments and Discussion}
\subsection{Details of Experiments}
We conducted image classification experiments on the CIFAR-10 dataset~\cite{2009-Report-Krizhevsky} and the Imagenette dataset~\cite{imagenette} which is a subset of 10 classes from ImageNet.
CIFAR-10 consists of 60,000 color images (with a dimension of $3 \times 32 \times 32$) with 10 classes (6000 images for each class) where 50,000 images are for training and 10,000 for testing.
Imagenette consists of 13,394 color images with a (70/30 --- training/validation) split.

We used an opensource PyTorch implementation of ViT and fine-tuned a ViT-B\_16 model that was pre-trained with the Imagenet21k dataset. The parameters were a batch size of 512, a learning rate of 0.03, and a momentum value of 0.9 on a stochastic gradient descent (SGD) optimizer with a linear warmup and cosine decay scheduler. The model was fine-tuned for 10,000 steps for plain images, and 20,000 steps for encrypted ones.

For ConvMixer, we used the timm training framework as in the original ConvMixer paper with settings: RandAugment, mixup, CutMix, and random erasing in addition to timm default augmentation. However, for encrypted images, we used only mixup and CutMix. A batch size of 64 was used, and we trained ConvMixer networks for 300 epochs for all datasets. For CIFAR-10, ConvMixer was configured with an embedding dimension ($D = 256$), number of repetitions of ConvMixer layers ($depth = 8$), patch size ($P = 1$), and kernel size ($k = 9$). For Imagenette, we used $D = 384$, $depth = 8$, $P = 4$, and $k = 9$. The optimizer, AdamW, was used with a triangular learning schedule with a learning rate of $0.01$ and a weight decay value of $0.01$ for all datasets.

\subsection{Classification Performance}
We denoted ConvMixer models with ConvMixer-$D/depth$ and ViT model with ViT-B\_16 (because a patch size value of 16 was used for the ViT).
We trained the models with plain or encrypted images in different settings.

Table~\ref{tab:results} summarizes the classification results \RB{with the number of parameters of each model} for both datasets compared with state-of-the-art learnable encryption methods such as LE~\cite{2018-ICCETW-Tanaka}, ELE~\cite{madono2020block}, and PE~\cite{2019-Access-Warit}. The ConvMixer models were extremely smaller (lower number of parameters). They were trained from scratch (without pre-training) and provided the highest accuracy for EtC images. \RB{In contrast, the ViT-B\_16 models were the biggest with more than 85 million parameters.} By taking advantage of a pre-trained model, ViT outperformed every other model in classifying plain images, but the accuracy dropped for EtC images. Besides, conventional methods were never tested for a larger image size, and the accuracy of classifying EtC images was lower compared with ConvMixer.

In addition, we visualized patch embeddings (256 filters) of ConvMixer-256/8 with $P = 1$ for CIFAR-10, and those of ConvMixer-384/8 with $P = 4$ (the first 256 filters) for Imagenette in Fig.~\ref{fig:weight-visualize}.
As in~\cite{anonymous2022patches}, we observed different patterns in patch embedding weights.
The figures show that there are high similarities between patch embeddings of plain images and encrypted ones.
This also gives us some insight on why isotropic networks are able to classify encrypted images with a high classification accuracy.

\subsection{Compressibility}
\RB{
Table~\ref{tab:jpeg} summarizes the classification performance of models trained and tested using lossy compressed JPEG images with varying quality factors. Compared with the conventional encryption, ELE~\cite{madono2020block}, the proposed models did not drop the accuracy significantly. Regarding lossless compression, it is expected to have the same classification performance as using uncompressed images, but the file size reduction is lower than with lossy compression.
}

\RB{
A file size analysis of the CIFAR-10 dataset is shown in Table~\ref{tab:filesize} for both lossy and lossless compression methods. All images from the dataset (60,000) were used to calculate the file size. EtC provided almost the same compression performance as using plain images, attaining a reduction in file size of more than \SI{50}{\percent} under the use of lossy compression. Therefore, the proposed privacy-preserving image classification achieves a high classification accuracy with an additional advantage of compressibility.
}

\robustify\bfseries
\sisetup{table-parse-only,detect-weight=true,detect-inline-weight=text,round-mode=places,round-precision=2}
\begin{table}
\centering
\caption{\RB{Classification accuracy (\SI{}{\percent}) of models trained and tested using compressed JPEG images with varying quality factors. ($QF$) denotes quality factor. ConvMixer-256/8, ViT-B\_16, and ShakeDrop were used for CIFAR-10 dataset, and ConvMixer-384/8 and $^\ast$ViT-B\_16 were used for Imagenette dataset.\label{tab:jpeg}}}
\resizebox{\columnwidth}{!}{%
\begin{tabular}{cSSS}
 \toprule
 {Model} & {(Uncompressed)} & {($QF = 85$)} & {($QF = 80$)}\\
 \midrule
 ConvMixer-256/8 (EtC) & 92.72 & 90.43 & 89.95\\
 ConvMixer-384/8 (EtC) & 90.11 & 89.17 & 88.91\\
 ViT-B\_16 (EtC) & 87.89 & 81.99 & 81.84\\
 $^\ast$ViT-B\_16 (EtC) & 90.62 & 89.63 & 89.60\\
 \midrule
 ShakeDrop (ELE~\cite{madono2020block}) & 83.06 & 31.5 & 30.51\\
 \bottomrule
\end{tabular}
}
\end{table}

\robustify\bfseries
\sisetup{table-parse-only,detect-weight=true,detect-inline-weight=text,round-mode=places,round-precision=2}
\begin{table}
\centering
\caption{\RB{File size analysis of two encryption methods with different compression settings for CIFAR-10 dataset. ($QF$) denotes quality factor. Values are in unit of megabyte (MB).\label{tab:filesize}}}
\begin{tabular}{cSSS}
 \toprule
 & {Plain} & {EtC~\cite{2019-TIFS-Chuman}} & {ELE~\cite{madono2020block}}\\
 \midrule
 {JPEG $QF = 85$} & 68.7 & 69.0 & 134.5\\
 {JPEG $QF = 80$} & 64.4 & 64.7 & 122.4\\
 \midrule
 {JPEG-LS} & 135.9 & 139.7 & 222.1\\
 {JPEG 2000} & 157.3 & 163.8 & 228.6\\
 \midrule
 {Uncompressed} & \multicolumn{3}{c}{178.20}\\
 \bottomrule
\end{tabular}
\end{table}

\begin{figure}[t]
\centerline{\includegraphics[width=18.5pc]{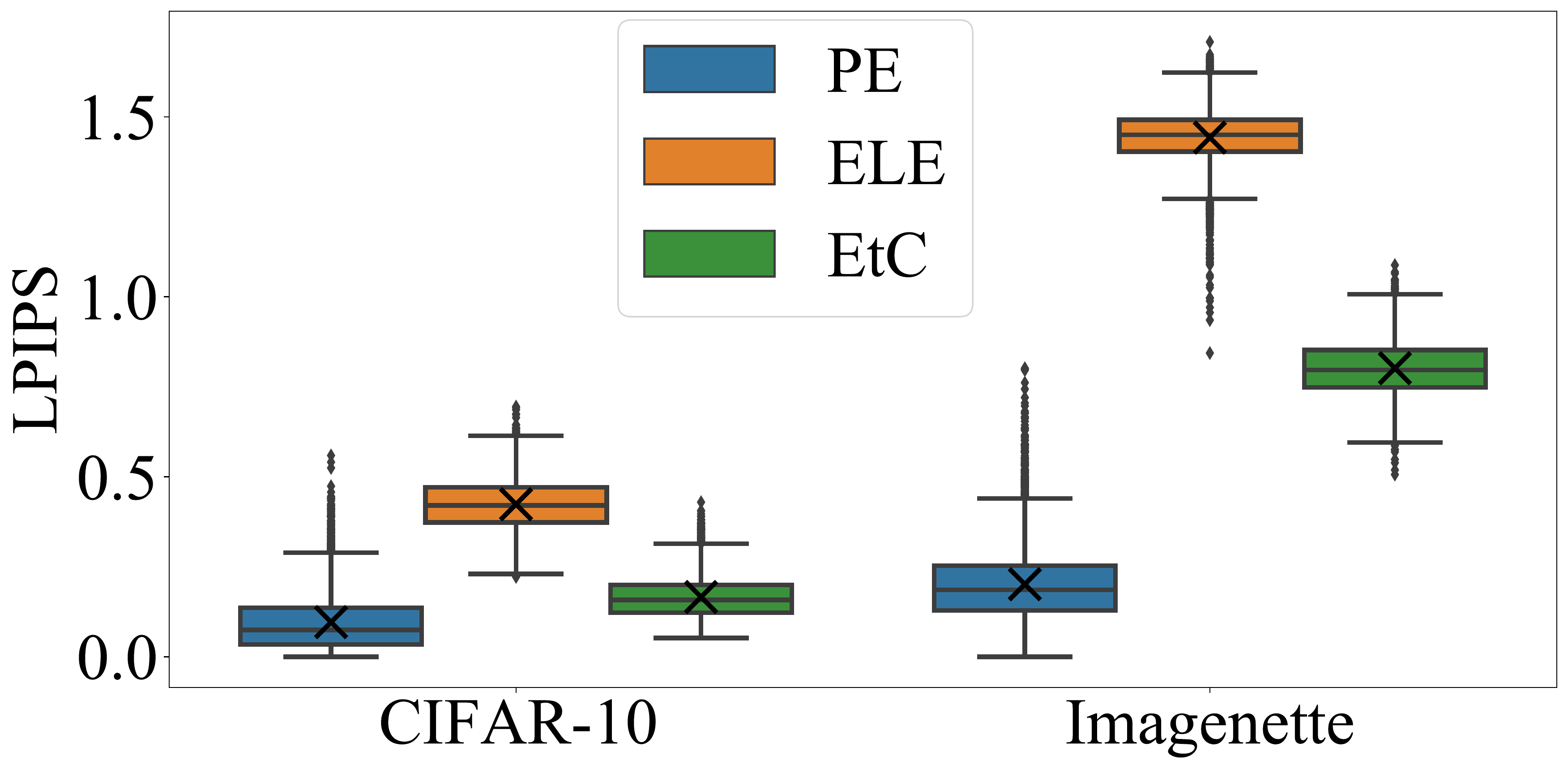}}
\caption{\RC{LPIPS scores of encrypted images (PE~\cite{2019-Access-Warit}, ELE~\cite{madono2020block}, EtC~\cite{2019-TIFS-Chuman}) for both CIFAR-10 and Imagenette datasets. Randomly chosen 3,856 images from test set were used to plot graph. Boxes span from first to third quartile, referred to as $Q_1$ and $Q_3$, and whiskers show maximum and minimum values in range of $[Q_1 - 1.5(Q_3 - Q_1), Q_3 + 1.5(Q_3 - Q_1)]$. Band and cross inside boxes indicate median and average values, respectively. Dots represent outliers.\label{fig:lpips}}}
\end{figure}

\begin{figure*}[t]
\centerline{\includegraphics[width=37pc]{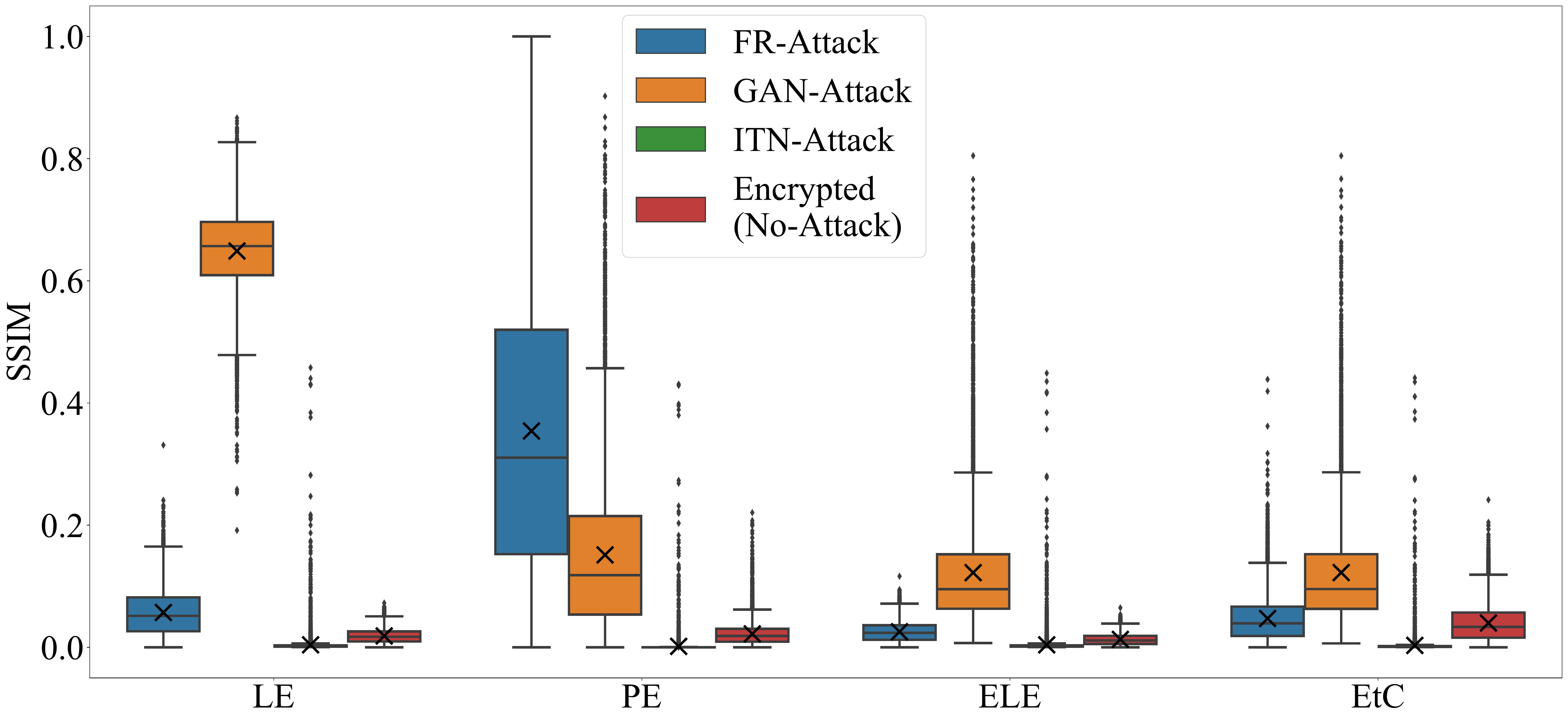}}
\caption{\RB{SSIM values of reconstructed images under various attacks for different encryption methods. Test set (10,000 images) from CIFAR-10 was used to plot graph. Boxes span from first to third quartile, referred to as $Q_1$ and $Q_3$, and whiskers show maximum and minimum values in range of $[Q_1 - 1.5(Q_3 - Q_1), Q_3 + 1.5(Q_3 - Q_1)]$. Band and cross inside boxes indicate median and average values, respectively. Dots represent outliers.\label{fig:ssim}}}
\end{figure*}

\subsection{Visual Information on Encrypted Images}
To measure visual information on encrypted images, we calculated Learned Perceptual Image Patch Similarity (LPIPS version=$0.1$)~\cite{zhang2018unreasonable} scores to quantify perceptual information on encrypted images. Figure~\ref{fig:lpips} shows LPIPS scores for three encryption schemes: PE~\cite{2019-Access-Warit}, ELE~\cite{madono2020block}, and EtC~\cite{2019-TIFS-Chuman} for both CIFAR-10 and Imagenette. Higher scores indicate encrypted images are not perceptually similar to plain images. \RC{In general, the scores for Imagenette were higher than those for CIFAR-10 because the image size of Imagenette ($224 \times 224$) was bigger than that of CIFAR-10 ($32 \times 32$). Although ELE provided the highest score, ELE images are not compressible and require an adaptation network.} From the LPIPS \RC{scores}, we confirmed that EtC images with a large size were not visually similar to their plain counterparts, suggesting visual information protection.

\subsection{Robustness Against Attacks}
We performed the FR-Attack~\cite{chang2020attacks}, GAN-Attack~\cite{madono2021gan}, and ITN-Attack~\cite{ito2021image} to evaluate the robustness of the EtC images. The results are summarized in Fig.~\ref{fig:sec-eval}. Some visual information was restored for encrypted images with LE~\cite{2018-ICCETW-Tanaka} and PE~\cite{2019-Access-Warit}. \RB{In addition, the structural similarity index measure (SSIM) values of the 10,000 attacked images (test set of CIFAR-10) are plotted in Fig.~\ref{fig:ssim}.} \RA{Although ELE~\cite{madono2020block} was robust against such attacks, it cannot be applied to images with a large size due to the network expansion in the adaptation network, and the encrypted images in ELE are not compressible.} In contrast, EtC images with a large image size are applicable in the proposed privacy-preserving image classification, and are not only robust against the above attacks, but also compressible by the JPEG standard.

\subsection{Limitations of Proposed Classification}
\RB{
EtC images are expected to meet three requirements: compressibility, no visual information on plain images, and a useful classification function that a network can learn. ViT and ConvMixer were shown to provide the best performance with the conventional methods with perceptual encryption in terms of the three requirements. However, the proposed method still has possible limitations the same as other perceptual-encryption-based methods. Perceptual-encryption-based methods do not have provable security, so they do not provide the same privacy/confidentiality as standard encryption, but they provide some advantages like the use of light-weight encryption and networks without special preparation. In the paper, EtC images were experimentally shown to be robust against ciphertext-only attacks (COA) in some settings.
In the above scenario, the keys used for the encryption need to securely be managed, and \RC{pairs of encrypted images and their decryption should not be disclosed.}
}

\section{Conclusion and Future Work}
In this paper, we proposed a privacy-preserving image classification method that uses EtC images and an isotropic network.
The proposed solution allows us to use a large image size and achieves high classification, robustness against attacks, and efficient compression.
The experiment results show that EtC with ConvMixer outperformed previous learnable image encryption methods without needing any adaptation network. As for future work, we shall investigate the aggregation of multiple datasets encrypted with different keys, improve the classification performance, and identify potential threats for further analysis.

\bibliographystyle{IEEEtran}
\bibliography{IEEEabrv,/Users/maung/Dropbox/refs}

\end{document}